\pdfoutput=1
\documentclass[journal]{IEEEtran}
\usepackage{graphicx} 
\usepackage{caption}
\usepackage{multirow}
\usepackage{booktabs}
\usepackage{bigstrut}
\usepackage{hyperref}
\usepackage[table]{xcolor}
\usepackage{amsmath}
\usepackage{setspace}
\usepackage{changes}
	\definechangesauthor[name={Per cusse}, color=orange]{per}
\ifCLASSINFOpdf
\else
\fi
\begin{document}
%

\title{Color-related Local Binary Pattern: A Learned Local Descriptor for Color Image Recognition}
%
%
%

\author{Bin Xiao,
        Tao Geng,
        Xiuli Bi,
        Weisheng Li
\thanks{This work was partly supported by the National Key Research and Development Project (2016YFC1000307-3), the National Natural Science Foundation of China (61806032 and 61976031), the Chongqing Research Program of Application Foundation ${\text{\& }}$ Advanced Technology (cstc2018jcyjAX0117) and the Scientific ${\text{\& }}$ Technological Key Research Program of Chongqing Municipal Education Commission (KJZD-K201800601). (Corresponding authors: Xiuli Bi, Weisheng Li)}
\thanks{ Bin Xiao, Tao Geng, Xiuli Bi and Weisheng Li are with the Department of Computer Science and Technology, Chongqing University of Posts and Telecommunications, Chongqing, China. email: xiaobin@cqupt.edu.cn, 13253603925@163.com, bixl@cqupt.edu.cn, liws@cqupt.edu.cn.}}
%
%

\markboth{Journal of \LaTeX\ Class Files,~Vol.~14, No.~8, August~20XX}%
{Shell \MakeLowercase{\textit{et al.}}: Bare Demo of IEEEtran.cls for IEEE Journals}
%



\maketitle

\begin{abstract}
Local binary pattern (LBP) as a kind of local feature has shown its simplicity, easy implementation and strong discriminating power in image recognition. Although some LBP variants are specifically investigated for color image recognition, the color information of images is not adequately considered and the curse of dimensionality in classification is easily caused in these methods. In this paper, a color-related local binary pattern (cLBP) which learns the dominant patterns from the decoded LBP is proposed for color images recognition. This paper first proposes a relative similarity space (RSS) that represents the color similarity between image channels for describing a color image. Then, the decoded LBP which can mine the correlation information between the LBP feature maps correspond to each color channel of RSS traditional RGB spaces, is employed for feature extraction. Finally, a feature learning strategy is employed to learn the dominant color-related patterns for reducing the dimension of feature vector and further improving the discriminatively of features. The theoretic analysis show that the proposed RSS can provide more discriminative information, and has higher noise robustness as well as higher illumination variation robustness than traditional RGB space. Experimental results on four groups, totally twelve public color image datasets show that the proposed method outperforms most of the LBP variants for color image recognition in terms of dimension of features, recognition accuracy under noise-free, noisy and illumination variation conditions.

\end{abstract}

\begin{IEEEkeywords}
Image classification, LBP, relative similarity space, color image, local descriptor.
\end{IEEEkeywords}

%
\IEEEpeerreviewmaketitle

\section{Introduction}
%
%
%
%
\IEEEPARstart{I}{mage} descriptors play a key role in many computer vision related applications such as: image retrieval \cite{arandjelovic2012three}, object and scene recognition \cite{lowe1999object}\cite{van2009evaluating}, image classification \cite{lin2014study}, etc. These applications primarily rely on extracting image features and analyzing these features to classify the image or obtain the image that is most similar to the target image. Therefore, a well designed descriptor can greatly improve the recognition performance and processing speed. In recent years, many descriptors have been proposed. These descriptors can be roughly divided into three categories: local features, global features and deep learning-based features \cite{zagoruyko2015learning}. Among these three kinds of descriptors, local features focus on the local information between pixels in the image so that the pattern matching will not be affected by the local deviation. This kind of descriptors has a great advantage in the computer vision related applications, and mainly include SURF \cite{bay2008speeded}, DAISY \cite{tola2009daisy}, BRIEF \cite{calonder2011brief}, etc.

Local binary pattern (LBP) as a kind of widely used local feature was proposed and optimized by Ojala \emph{et al.} \cite{ojala2000gray}\cite{ojala2002multiresolution}. It has received extensive attention because of its simplicity, easy implementation and strong discriminating power. Based on the traditional LBP, many variants of LBP have been introduced. Some variants change the coding and mode selection strategies such as LBP-TOP \cite{zhao2007dynamic}, LTP \cite{tan2010enhanced}, disLBP \cite{guo2012discriminative}, PRICoLBP \cite{qi2014pairwise}, 2D-LBP\cite{xiao20182d}, rotation-invariant local binary descriptor (RI-LBD) \cite{Duan2017}, etc. Some variants change the neighborhood topologies and sampling structures such as CSLBP \cite{heikkila2009description}, ELBP \cite{nguyen2012elliptical}, median binary pattern\cite{Alkhatib2019}, etc. As the most prominent information in an image, the color information is closely related to objects or scenes, which is taken as one of the most widely used feature in image recognition and retrieval. However, most of the LBP variants focus on the texture information and the color information is always ignored. The color images are usually converted to gray-scale images firstly when processing color images by these variants of LBP \cite{ojala2002multiresolution}.

\begin{figure*}[hbt]
\centering
\includegraphics[width=\textwidth]{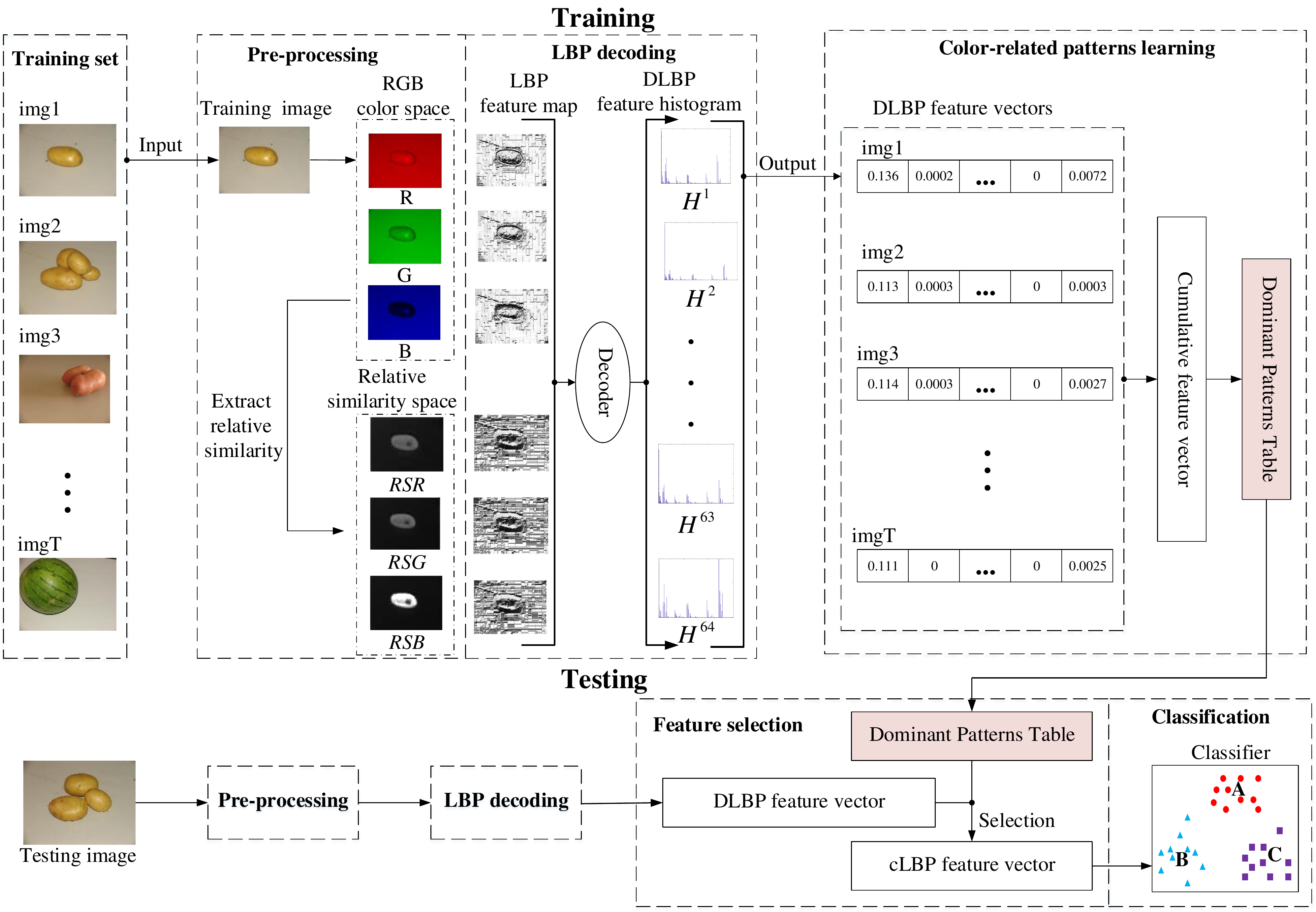}
\caption{The framework of the proposed cLBP for color image representation.}
\label{fig_sim}
\end{figure*}

To utilize the color information, recently, some variants of LBP have been proposed for color image recognition. Zhu \emph{et al.} proposed the orthogonal combination of local binary pattern (OC-LBP) \cite{zhu2013image} and a new local descriptors based on OC-LBP enhanced with color information for image description. In color radial mean completed local binary pattern (CRMCLBP) \cite{sotoodeh2019novel}, the radial mean completed local binary pattern \cite{shakoor2018radial} was computed on the color channels independently. To consider the correlation information between color channels of an image, multispectral local binary pattern (MSLBP) \cite{maenpaa2002separating} that used the opponent LBP to capture the spatial relationship between two color channels was proposed for describing a color image by six sets of opponent LBPs and three LBPs computed from each spectrum independently. Lee \emph{et al.} \cite{lee2011local} proposed the local color vector binary pattern (LCVBP) to extract the characteristics of color images by color norm patterns and color angular patterns, where the color angular patterns were calculated by the ratio among different spectral-band images. Lan \emph{et al.} \cite{lan2015quaternionic} proposed a quaternion local ranking binary pattern (QLRBP), which represents the image by quaternion algebra. The QLRBP operator calculated the similarity between each pixel and the reference point in the image by the phase of Clifford translation of quaternion. Li \emph{et al.} \cite{li2016completed} proposed the completed local similarity pattern (CLSP) for color image recognition that was consisted of two parts: color labeling, and local similarity pattern which calculates LBP from the color distance between the central pixel and the neighborhood pixels. Singh \emph{et al.} \cite{singh2018color} developed a color texture descriptor called Local binary pattern for color images (LBPC). LBPC divided the neighbor of central pixel into two categories by establishing a spatial threshold plane, and can be fused with the local binary pattern of the Hue channel and color histogram to boost the discriminative power. Dubey \emph{et al.} \cite{dubey2016multichannel} proposed a method to extract color information called multichannel decoded local binary pattern (mdLBP), which combines the information from the R, G and B color channels in a decoding manner. In the mdLBP, two schemas, i.e., adder and decoder were used for capturing the joint information of multichannel. Compared with other LBP descriptors, mdLBP combines the joint information between each color channel well and keeps the primary information of color channels.
\\
\indent In summary, most of existing LBP variants for representing color images cannot adequately consider the color information and contain much redundant information, which results in increasing the dimension of feature vector and decreasing the recognition performance. In this paper, we propose a local color descriptor named color-related local binary pattern (cLBP) which learns the dominant color-related patterns from the decoded LBP for color images representing. In the proposed method, the relative similarity space (RSS) is firstly proposed to obtain the color similarity between the three channels of color images. Then, the LBP decoding is employed to describe the color image on the combination of the RSS and traditional RGB color spaces. Finally, a feature learning strategy is used to learn the most discriminative features (dominant color-related patterns) and reduce the dimension of feature vector. Experimental results show that the proposed cLBP can achieve a promising result on texture, object and face recognition under noisy, noise-free and illumination variation conditions.

The rest of this paper is organized as follows. The framework of color image recognition by the proposed cLBP is introduced in Section 2. In Section 3, the effectiveness of the proposed cLBP is demonstrated through experiments. Finally, conclusions are remarked in Section 4.


\section{The proposed cLBP}
\indent
This section first introduces the framework of color image recognition by the proposed cLBP. Then, the RSS of color image is provided in subsection A. In subsection B, the scheme of LBP decoding on multi-color channels is introduced. The color-related patterns learning on the decoded LBP features is described in subsection C.

The framework of the proposed cLBP for color image recognition is shown in Fig. 1. It mainly contains two procedures: the training and testing procedures. The significance of training procedure is to obtain an adaptive dominant pattern table by learning dominant patterns from the decoded LBP of training images. The training procedure is consisted of three stages: pre-processing, LBP decoding, and color-related patterns learning. For a training image, to well describe the color information, in the pre-processing stage, the RSS and traditional RGB color spaces are combined to fully represent the color information. On this basis, in the LBP decoding stage, the traditional LBP operation is performed on the six channels to obtain the corresponding LBP feature maps respectively. This is followed by decoding these LBP feature maps to capture the joint color information of LBP feature maps, which corresponds to each color channel. Since totally six color channels are used, there are 64 histograms outputted by the decoded LBP, and each of which consists of 256 bins. These decoded LBP feature histograms are concatenated to form a feature vector for color image representing. To remove the redundancy of the obtained feature vector, a color-related pattern learning strategy is applied on the decoded LBP to improve the recognition accuracy and efficiency. In the color-related pattern learning stage, the cumulative histogram is calculated by adding the decoded LBP feature vectors of all the training images, and the dominant pattern table is obtained from the cumulative histogram by feature selecting strategy, which will be provided in detail in subsection II.C. Therefore, in the testing procedure, the cLBP of color image is obtained by selecting the dominant color-related patterns from its decoded LBP according to the learned dominant pattern table, and finally, this is followed by feeding into a classifier for image recognition.

\subsection{The Relative Similarity Space (RSS)}
 In this subsection, the relative similarity space (RSS) is proposed to represent the color similarity between the R, G and B channels of a color image. The relative similarity considers the joint distribution of each channel of a color image, which can well represent the cross-similarity between color channels. By the proposed RSS, more color information is considered for the following feature extraction and learning stages. Moreover, since the calculation process in RSS can offset the interference of noise and illumination, it makes the proposed cLBP performs well in color image recognition under both noisy and illumination variation conditions.

Unlike gray-scale image, the color image consists of multiple channels. Therefore, the similarity is considered to represent the relationship between color channels. For two positive numbers $p$ and $q$ (the values of RGB channels are ranged in $[0, 255]$), the similarity of them can be calculated by their difference as
\begin{equation} \label {eq1}
S = {\kern 1pt} {\kern 1pt} |p - q|
\end{equation}
In this case, if these two numbers are similar to each other, the similarity $S$ will tend to 0. However, Eq. \eqref{eq1} ignores the order of these two numbers and the value of these numbers themselves. For example, $(p = 1,{\kern 1pt} {\kern 1pt} q = 2)$ and $(p = 2,{\kern 1pt} {\kern 1pt} q = 1)$ have the same similarity by Eq. \eqref{eq1}. The difference between 1 and 2 is the same as the difference between 100 and 101, but these two sets of numbers are at different level of value. To avoid these problems, the relative similarity is considered. The similarity of $p$ relative to $q$ is defined as
\begin{equation}  \label {eq2}
RS = \frac{{|p - q|}}{{p + \xi }}
\end{equation}
where $\xi $ is a small value to avoid the denominator being 0. The change of relative similarity $RS$ with $p$ and $q$ is shown in Fig. 2. From this figure, it can be found that relative similarity is always 0 when the values of the two parameters are equal. The relative similarity increases as the difference between the two numbers increases. The relative similarity decrease with the increase of number value when the difference between the two number is a constant. Most importantly, there are two different increment rates of the relative similarity. When $p > q$, the relative similarity increases slowly as the difference between the two numbers increases, and is always less than 1. On the contrary, the relative similarity increases rapidly when $p < q$. This phenomenon means that the relative similarities are different when two numbers have different orders.
\begin{figure}[!t]
\centering
\includegraphics[width=3in]{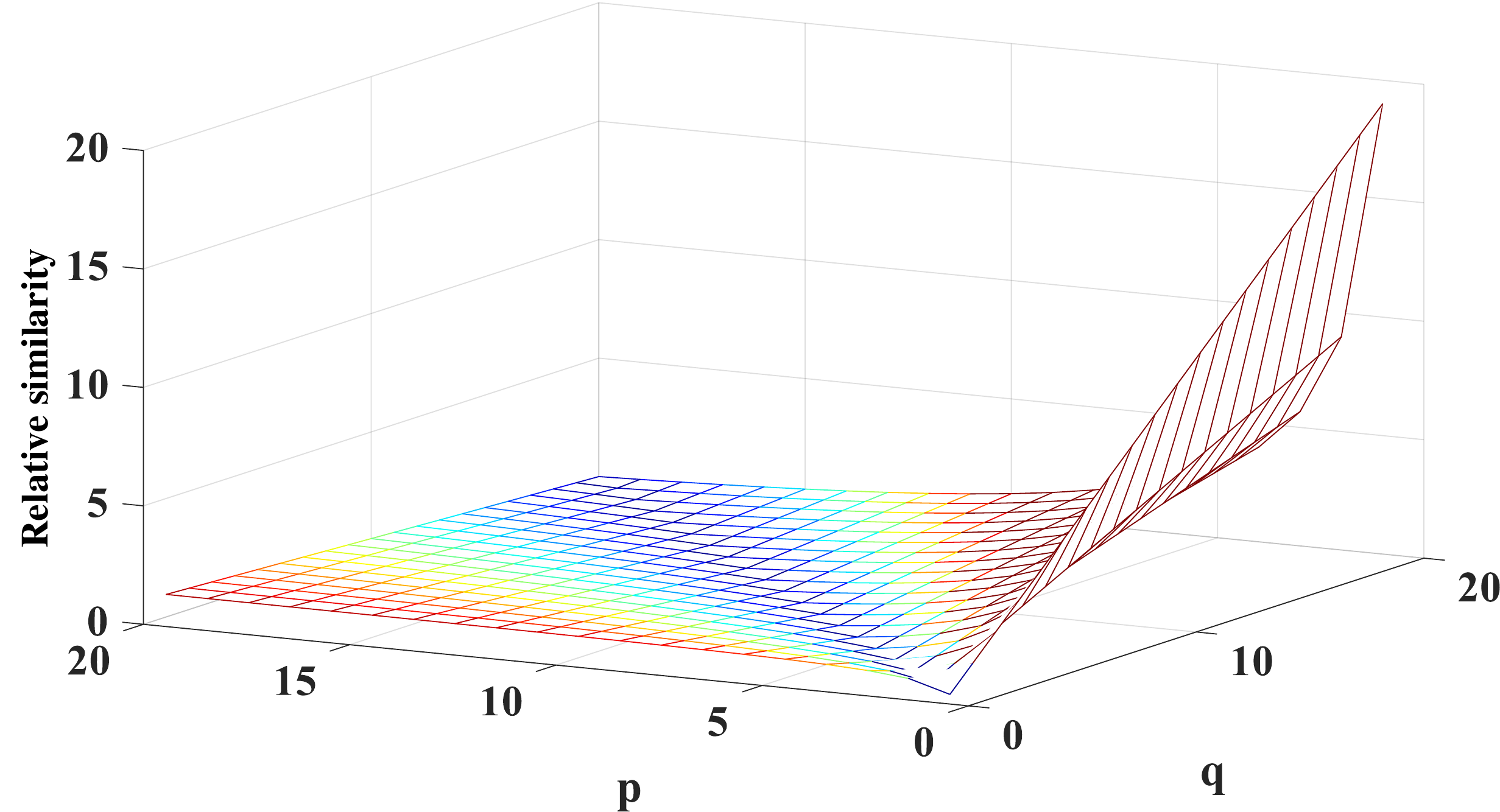}
\caption{The change of relative similarity $RS$ with the change of $p$ and $q$.}
\label{fig_sim}
\end{figure}
\begin{figure}[!t]
\centering
\includegraphics[width=3in]{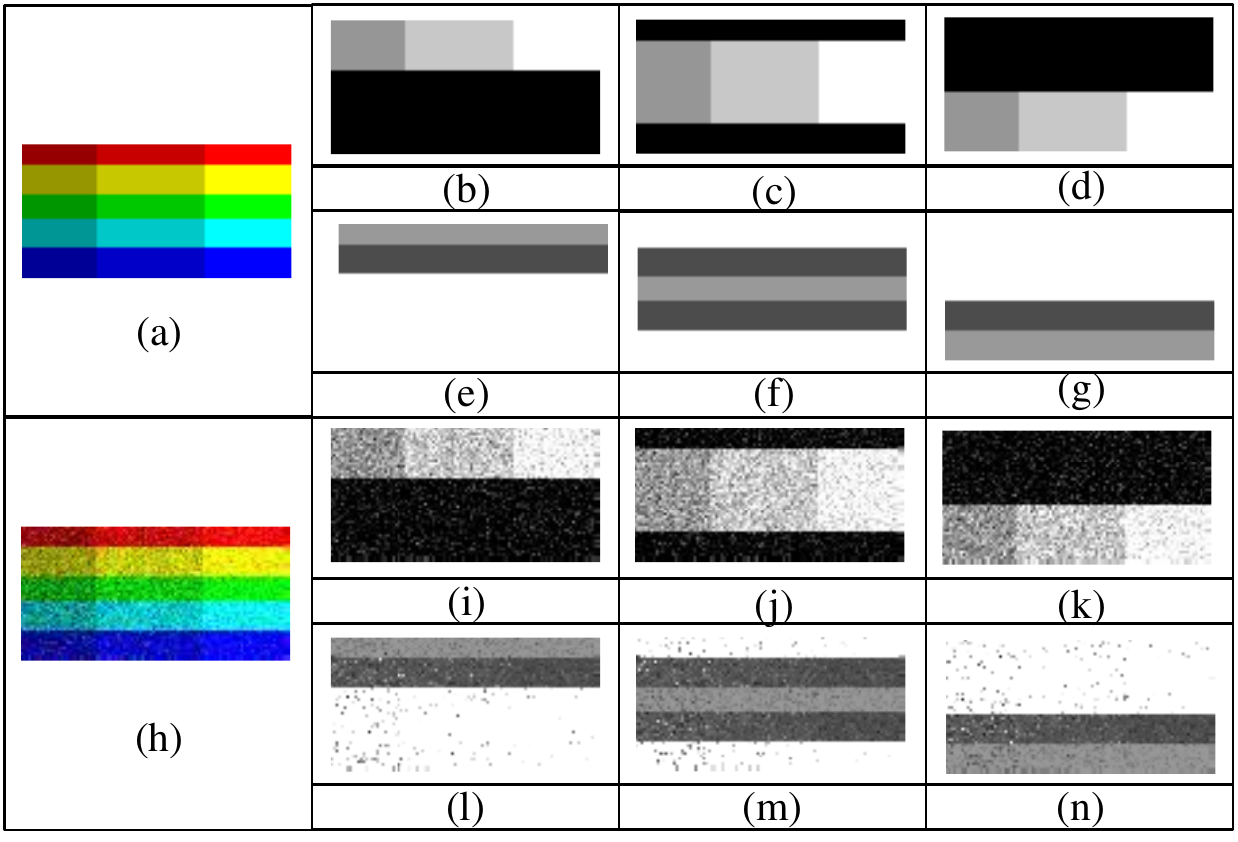}
\caption{An example of color image (with illumination variation) represented in the RGB and RSS space under both noisy and noise-free conditions. (a) original color image, (b) R channel image, (c) G channel image, (d) B channel image, (e) $RSR$ channel image, (f) $RSG$ channel image, (g) $RSB$ channel image, (h) color image under noise condition, (i-k) R, G and B images under noise condition, (l-n) RSR, RSG and RSB images under noise condition.}
\label{fig_sim}
\end{figure}
\begin{figure*}[!ht]
\centering
\includegraphics[width=0.8\textwidth]{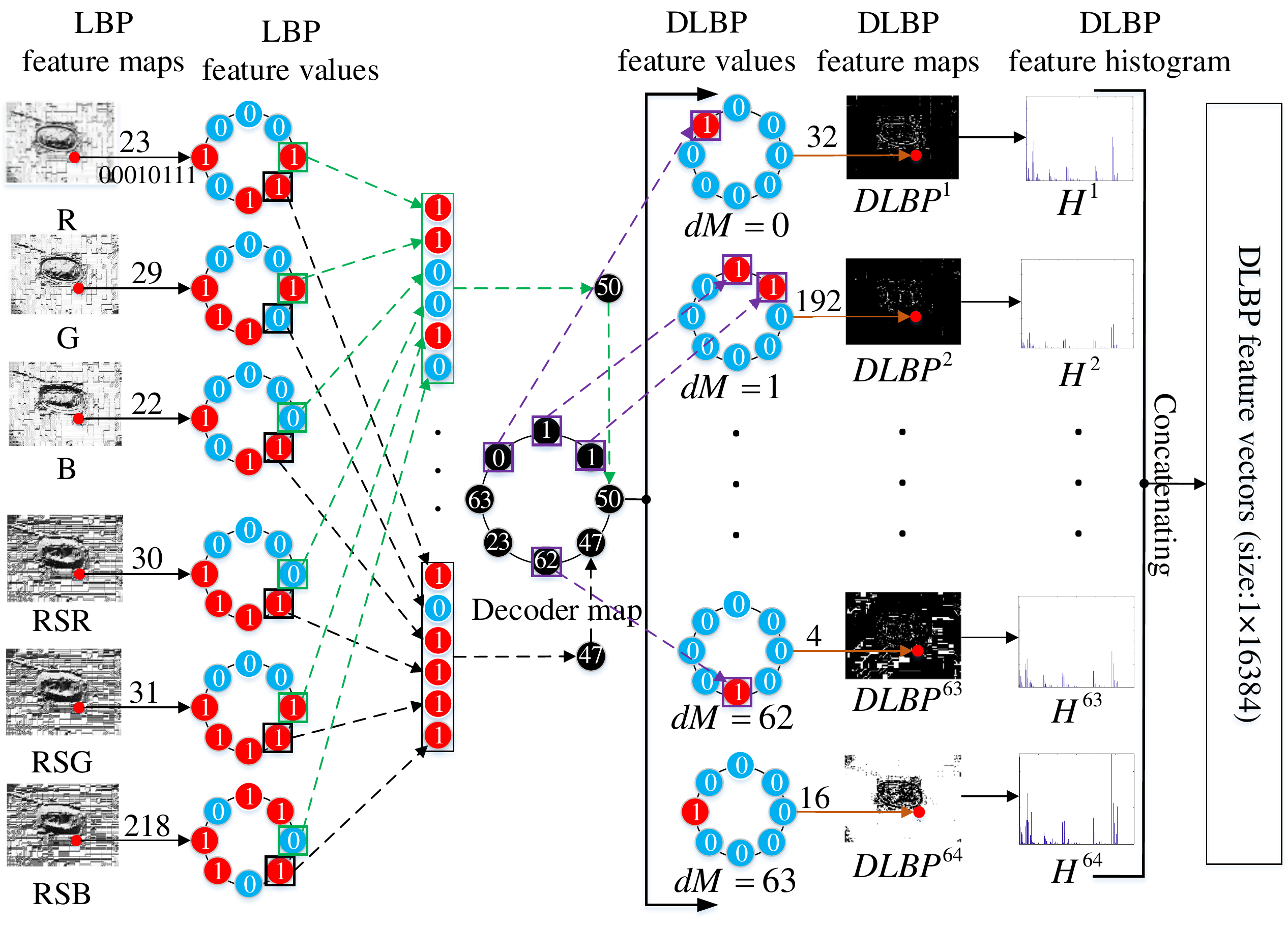}
\caption{Scheme of LBP decoding on the combination of RSS and RGB color spaces with $P$ = 8 .}
\label{fig_sim}
\end{figure*}

Based on Eq. \eqref{eq2}, the relative similarity can be applied for representing the color images. Let $R(x,y)$, $G(x,y)$ and $B(x,y)$ represent the values of color channels correspond to a pixel located at $(x,y)$, the relative similarity of G and B related to R can be computed as
\begin{equation}
RS{R(x,y)} = \frac{{|R(x,y) - G(x,y)| + |R(x,y) - B(x,y)|}}{{R(x,y) + \xi }}
\end{equation}
Similarly, the relative similarity of R and B related to G, R and G related to B are defined as follows:
\begin{equation}
RS{G(x,y)} = \frac{{|G(x,y) - R(x,y)| + |G(x,y) - B(x,y)|}}{{G(x,y) + \xi }}
\end{equation}
\begin{equation}
RS{B(x,y)} = \frac{{|B(x,y) - R(x,y)| + |B(x,y) - G(x,y)|}}{{B(x,y) + \xi }}
\end{equation}

$RSR$, $RSG$ and $RSB$ represent the relative similarity of three color channels. They can be constructed as a new color space termed as RSS, which mainly describes the color similarity between the channels of a color image. Fig. 3 illustrates a color image (with illumination variation) and its three channels in RGB and RSS spaces under both noisy and noise-free conditions. We can find that these images in RSS space reflect the color similarity of the original color image well and is difficult to be affected by noise and illumination variation.
\subsection{LBP decoding on multi-color channels}
Fig. 4 illustrates the scheme of LBP decoding on the combination of RSS and RGB color spaces. It mainly consists of three steps. In the first step, after the color image is represented in RSS and RGB color space, the traditional LBP is used for encoding the color information of each channel respectively, and totally six LBP feature maps are obtained. Secondly, the LBP decoding step is followed to capture the joint information of these six LBP feature maps, by mapping the values in the LBP feature maps into the decoded LBP feature maps in a decoding manner \cite{dubey2016multichannel}.  At last, a set of decoded LBP feature histograms is generated, and this is followed by concatenating these histograms to form a feature vector to describe the color image.
\\
\indent For a color image of size $N\times M$ represented by RSS and RGB spaces $I(x,y) = (V_1(x,y), V_2(x,y),...,V_6(x,y)), x\in[1,N], y\in [1,M]$, the traditional LBP operated on the ${n^{th}} $ ($\forall n\in[1,6]$) channel is computed as
\begin{equation}  \label {eq6}
LBP_{n,R,P}(x,y) = \mathop \sum \limits_{m = 0}^{P - 1} S\left( {V_{n,R,P}^m(x,y) - {V_n}(x,y)} \right) \times {2^m}
\end{equation}
\begin{equation}
S\left( {\rm{x}} \right) = \left\{ {\begin{array}{*{20}{c}}
{1,\;\;\;\;x \ge 0}\\
{0,\;\;\;\;x < 0}
\end{array}} \right.
\end{equation}
where ${V_n}(x,y)$ is the value of the center pixel located at $(x,y)$  in the ${n^{th}}$ channel, and $V_{n,R,P}^m(x,y)$ represents the value of the ${m^{th}}$ neighboring pixel centered at $(x,y)$. $P$ is the total number of involved neighborhood pixels and $R$ is the radius of the neighborhoods around the center pixel.
\begin{figure*}[!ht]
\centering
\includegraphics[width=0.9\textwidth]{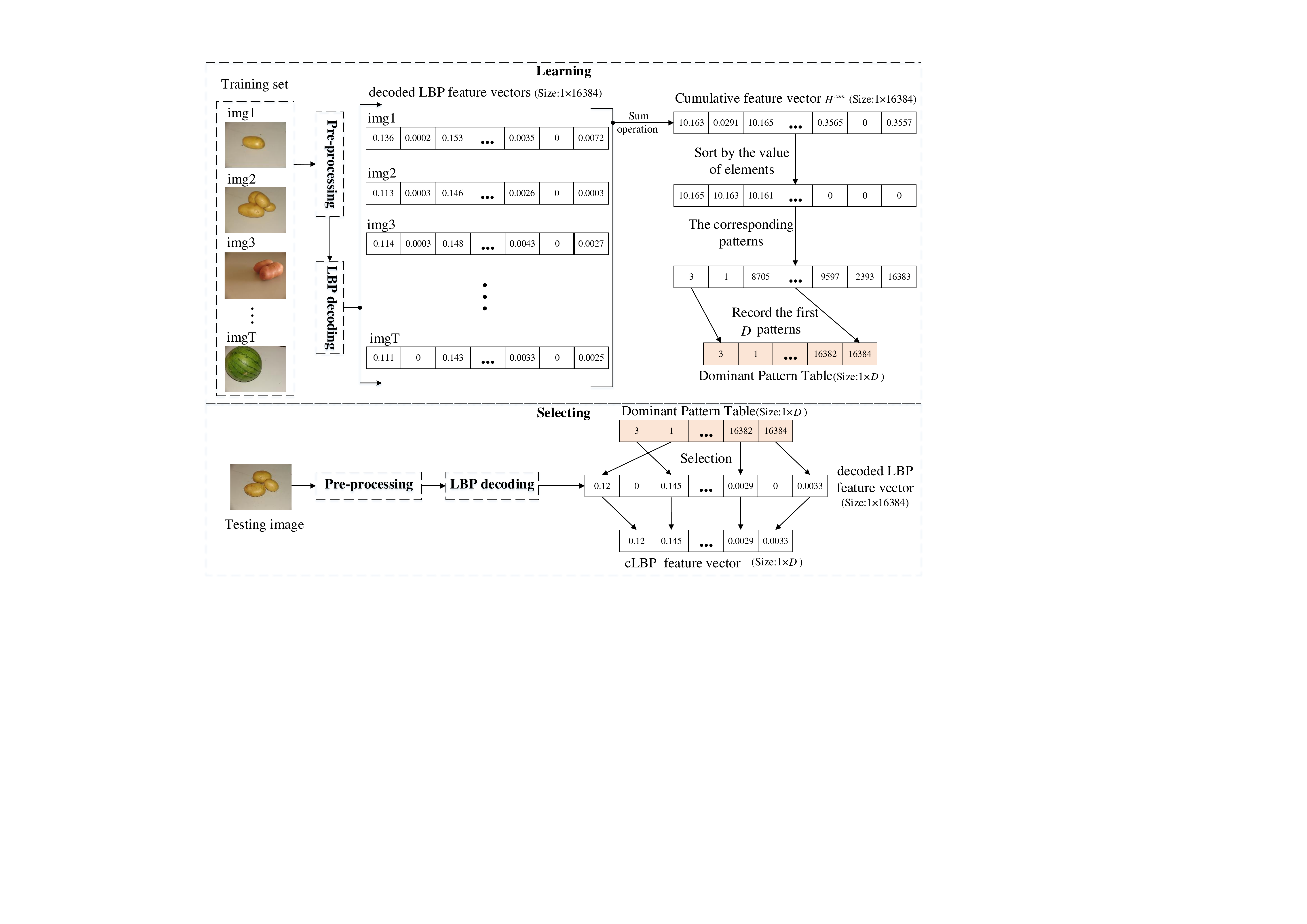}
\caption{An illustration of the color-related patterns learning and selecting procedures.}
\label{fig_sim}
\end{figure*}
\\
\indent Through Eq. \eqref{eq6}, there will generate six traditional LBP feature maps for a color image, and the values of these LBP feature maps are expressed as binary strings and ranged in $[0, 255]$. In the followed LBP decoding step, for the pixel located at $(x,y)$, six binary strings are corresponded in these feature maps. The ${m^{th}}$ bits of these six binary strings are concatenated to form a new binary string and denote as $d{M^m}(x,y)$  in the decoded LBP feature map. The $d{M^m}(x,y)$ can be mathematically expressed as
\begin{equation}
d{M^m}(x,y) = \sum\limits_{n = 1}^6 {{2^{(6 - n)}} \times S\left( {V_{n,R,P}^m(x,y) - {V_n}(x,y)} \right)}
\end{equation}
There will have 64 feature maps generated by the decoded LBP. The point $(x,y)$ in the ${c^{th}}$ ($\forall c \in [1,{2^6}]$) decoded LBP feature map $DLBP_c^{}(x,y)$ can be represented as
\begin{equation}
DLBP_c^{}(x,y) = \sum\limits_{m = 1}^P {DLBP_c^m(x,y) \times {2^{m - 1}}}
\end{equation}
where
\begin{equation}
DLBP_c^m(x,y) = \left\{ \begin{array}{l}
1,\;\;\;\;if\;d{M^m}(x,y) = (c - 1)\\
0,\;\;\;\;otherwise
\end{array} \right.
\end{equation}
Then, these decoded LBP feature maps are counted as feature histograms. The ${c^{th}}$ feature histogram ${H_c}$ can be computed as follows
\begin{equation}
{H_c}(k)=\frac{1}{{(N-2R)(M-2R)}}\sum\limits_{x = R + 1}^{N-R} {\sum\limits_{y = R + 1}^{M-R} {\delta (DLB{P_c}(x,y), k)} }
\end{equation}
\begin{equation}
\delta (u,v) = \left\{ \begin{array}{l}
1,\;\;\;\;if\;u = v\\
0,\;\;\;otherwise
\end{array} \right.
\end{equation}
where $\forall k \in [0,{2^P} - 1]$. Finally, by concatenating these feature histograms in a series, the feature vector of decoded LBP is provided as
\begin{equation} \label {eq13}
DLBP = \frac{1}{{{2^6}}}[{H_1},{H_2},...,{H_{{2^6}}}]
\end{equation}
\indent Compared with simply concatenating the LBP feature histograms from different color channels, the decoded LBP is an excellent way to mine the correlation information between the LBP feature maps corresponds to each color channel. However, from Eq. \eqref{eq13}, we can find that, the dimension of decoded LBP feature vector is almost ten times ($64\times 256$ to $6 \times 256$) larger than that of simply concatenating. In the next subsection, a simple but effective feature learning scheme is introduced to learn the dominant color-related patterns based on the feature vector obtained by the decoded LBP.
\subsection{Color-related Patterns Learning}
The dimension of decoded LBP feature vector on the combination of RSS and RGB color spaces is $256\times 64=16384$, to avoid the curse of dimensionality in classification, we introduce a feature learning strategy to reduce the dimension and capture the discriminative features, i.e., the color-related patterns. In the dominant LBP \cite{liao2009dominant}, the authors mentioned that dominant patterns are the patterns with high frequency in the feature maps and represent image's primary information. The proposed color-related patterns learning strategy are shown in Fig. 5. As shown in this figure, in the learning procedure, the patterns with high frequency of occurrence in the decoded LBP feature vectors across all the training images are selected as the discriminative features. For all the training images, the corresponding decoded LBP feature vectors are added together to obtain a cumulative feature vector ${H^{cum}}$\replaced{,}{} which can be computed as follows
\begin{equation}
{H^{cum}} = \sum\limits_{t = 1}^T {DLBP^t}
\end{equation}
where $T$ is the number of training images. After the cumulative feature vector ${H^{cum}}$ is obtained, it is sorted by the value of elements, and those elements correspond to the first $D$ ($D$ is the number of dominant patterns we want to be learned) highest value in ${H^{cum}}$ are considered as the dominant color-related patterns of the training set. Finally, the positions of these elements are recorded as the dominant pattern table. In the selecting procedure, for an input image, we firstly calculate the decoded LBP feature vector through the pre-processing and LBP decoding stages. Then, the $D$ patterns in the decoded LBP feature vector are selected by the learned dominant pattern table. Finally, the feature vector of cLBP is obtained by concatenating these selected patterns as
\begin{equation}
cLBP = [{B^1},{B^2},...,{B^D}]
\end{equation}
where $B^i$ is the selected pattern. The patterns in cLBP are learned from the content and color information of training images. Therefore, the proposed cLBP can reduce the dimension of feature vectors and improve the recognition accuracy in color image recognition.
\begin{table*}  \footnotesize
  \centering
  \caption{The basic information of twelve color image datasets used in the experiments}
  \begin{spacing}{1.1}
    \begin{tabular}{|c|p{5em}<{\centering}|m{27em}|c|c|c|c|}
    \hline
    \multicolumn{1}{|c|}{\multirow{2}[2]{*}{Type}} & \multirow{2}[2]{*}{Name} & \multirow{2}[2]{*}{Descriptions} & \multirow{2}[2]{*}{Image Size} & \multicolumn{1}{c|}{\multirow{2}[2]{4em}{Number of Class}} & \multicolumn{1}{c|}{\multirow{2}[2]{4em}{Samples per class}} & \multicolumn{1}{c|}{\multirow{2}[2]{4em}{Images (Total)}} \\
          & \multicolumn{1}{c|}{} & \multicolumn{1}{c|}{} & \multicolumn{1}{c|}{} &       &       &  \\
    \hline
    \multicolumn{1}{|c|}{\multirow{3}[20]{*}{Texture}} & KTH-TIPS & This is a color texture dataset consisting of 10 kinds of textures, and each texture is captured by 9 different ratios, 3 different poses and 3 different lighting &200$\times$200  & 10    & 81    & 810 \\
 \cline{2-7}          & STex-512-splitted & This color texture dataset was obtained by dividing the images in the Salzburg Texture Image Database (STex) into 16 non-overlapping blocks. &128$\times$128 & 31    & {Vary} & 7616 \\
 \cline{2-7}          & Colored Brodatz & It is a color texture image dataset based on the Brodatz dataset. This dataset was pre-processed in the experiment so that each color image was divided into 256 non-overlapping images of the same size. & 40$\times$40 & 112   & 256   & 28672 \\
    \hline
    \multicolumn{1}{|c|}{\multirow{4}[20]{*}{Object}} & Wang or SIMPLIcity & This dataset is a subset of the Corel image library, which is a relatively rich image library collected by Corel.  & 256$\times$384 & 10    & 100   & 1000 \\
 \cline{2-7}          & Corel-10k & Similar to the Wang dataset, Corel-10k is also part of the Corel image library. However, this dataset is more complicated. & 192$\times$128 & 100   & 100   & 10000 \\
 \cline{2-7}          & FTVL  & The dataset is a color image dataset describing tropical fruits and vegetables. & 1024$\times$768 & 15    & {Vary} & 2633 \\
 \cline{2-7}          & Coil-100 & The dataset contains 100 objects with a wide variety of complex geometric. The images of each object are taken as the object is rotated on a turntable. & 128$\times$128 & 100    & 72   & 7200 \\
    \hline
    \multicolumn{1}{|c|}{\multirow{2}[20]{*}{Face}} & Color FERET & This dataset contains a total of 11338 color facial images with different expression, angle, and illumination. 1378 images were selected that only two eyes could be reliably identified for normalization\replaced{}{.}, these images included five pose angles from $-45^{\circ}$ to $45^{\circ}$.  & 768$\times$512 & 107   & {Vary} & 1378 \\
 \cline{2-7}          & AR face & The dataset contains over 4000 color facial images of 126 people with different expressions, illumination, and occlusion. The images with occlusions are excluded and 100 commonly used subjects (50 men and 50 women) were selected. & 120$\times$165 & 100   & 14    & 1400 \\
    \hline
    \multicolumn{1}{|c|}{\multirow{2}[28]{*}{\shortstack{Illumination\\variation}}} & Outex-14 & This dataset contains 4080 color images of 68 textures obtained under three illumination conditions.  & 128$\times$128 & 68   & 60 & 4080 \\
    \cline{2-7}          & ALOI & The dataset  contains 1000 small objects, each of which has been imaged under 12 illumination conditions. & 384$\times$288 & 1000   & 12    & 12000 \\
    \cline{2-7}          & CUReT & The images in CUReT dataset are recorded by 61 materials under 205 different viewing and illumination conditions. For each material, 92 images with enough region of texture can be visible are chosen.Then crop a central $200\times 200$ region from each image, and discard the rest of the background. & 200$\times$200 & 61   & 92    & 5612 \\
    \hline
    \end{tabular}%
  \end{spacing}
  \label{tab:addlabel}%
\end{table*}%

\section{Experimental Results and Analysis}
To verify the effectiveness of the proposed method, four groups of experiments are designed. The experimental setting and color mage datasets used for validation are introduced in subsection A and B. In subsection C, the color image recognition ability affected by the dimension of feature vector by the proposed cLBP is discussed. The comparisons with some state of the art LBP variants specifically designed for color image recognition are provided in subsection D. Subsection E analyses the noise robustness of the proposed cLBP in color image recognition. Subsection F validates the color image recognition ability of the proposed cLBP under illumination variation.

\subsection{Experimental Setting}
In our experiments, totally six state of the art LBP variants specifically designed for color image recognition, i.e., LBPC (2018, \cite{singh2018color}), mdLBP (2016, \cite{dubey2016multichannel}), QLRBP (2015, \cite{lan2015quaternionic}), RGB-OC-LBP (2013, \cite{zhu2013image}), LCVBP (2011, \cite{lee2011local}) and LBP of RGB (LBP-RGB, 2002, \cite{ojala2002multiresolution}) are used for comparison. In LBP-RGB, the traditional LBP operator is performed on the R, G and B channels of a color image respectively, and the feature histograms of these three LBP feature maps are concatenated for recognition. For LBPC, the plane normal is set as the values given by the authors, i.e., local average normal, and the reference point is set as the intensity value of the center pixels. Since LBPC can be combined with the LBP of the hue and color histogram to improve recognition accuracy, in our experiment, the combination of LBPC with hue and color histogram which shows the best performance in color image recognition in \cite{singh2018color} is chosen for comparison. For QLRBP, according to the suggestion from the authors in \cite{lan2015quaternionic}, three weight parameters ${\alpha _1}, {\alpha _2}, {\alpha _3}$ are set to 1, and the value of three reference points are set as $(1-{\varepsilon _{11}}, {\varepsilon _{12}}, {\varepsilon _{13}})$, $({\varepsilon _{21}}, 1-{\varepsilon _{22}}, {\varepsilon _{23}})$ and $({\varepsilon _{31}}, {\varepsilon _{32}}, 1-{\varepsilon _{33}})$, where ${\varepsilon _{mn}} \in \left[ {0,0.1} \right]$. For all of these methods, the parameters $P$ and $R$ are fixed to 8 and 1. Except for QLRBP and LCVBP, the source codes of other LBP variants are provided by the
corresponding authors and the default parameters provided by the authors are adopted to keep consistency with the results given in the original papers. The linear multi-class support vector machines classifier of ``LIBLINEAR" with default parameters is utilized for classification, and the 10-fold cross-validation is used to obtain the final classification accuracy.

\subsection{Experimental Datasets}
To thoroughly test the performance of the proposed cLBP for color image recognition, totally twelve public color datasets that can be divided into four groups: 1) KTH-TIPS \cite{fritz2004kth}, STex-512S \cite{stex} and Colored Brodatz \cite{abdelmounaime2013new} which belong to the color texture datasets; 2) Wang or SIMPLIcity \cite{wang2001simplicity}, Corel-10k \cite{liu2015content}, FTVL \cite{ftvl} and Coil-100 \cite{coil} which belong to the color object datasets; 3) Color FERET \cite{phillips2000feret} and AR face \cite{martinez1998ar} which belong to the color face datasets were utilized for validating the recognition accuracy and noise robustness. Moreover, to verify the color image recognition ability of the proposed method under illumination variation condition, the Outex-14 \cite{ojala2002outex}, ALOI \cite{geusebroek2005amsterdam} and CUReT \cite{dana1999reflectance} which have been widely used in existing methods, are also employed for validation in this paper. The detail description about these twelve color image datasets is summarized in Table \uppercase\expandafter{\romannumeral1}.

\subsection{Discussion on the Dimension of Features}
In this experiment, the classification accuracy of the proposed cLBP with different dimension of feature, is evaluated on three of the above twelve image datasets (from the texture, object and face groups), to study the effectiveness of the proposed pattern learning framework. The cLBP only extracted from the RGB space (cLBP-RGB), RSS space (cLBP-RSS) and the combination of RGB and RSS spaces (cLBP) are used for comparison. Since the dimension of cLBP-RGB and cLBP-RSS is 2048 (without patterns learning step), the dimension of learned dominant color-related patterns in cLBP (i.e., $D$) is varied from 100 to 2000 with 100 increments. The experimental results are shown in Fig. 6. It can be seen from this figure that, for both cLBP-RGB and cLBP-RSS, the recognition accuracy increase firstly and then decreases with the increment of feature dimension especially for the ``Corel-FERET" dataset, while for the cLBP, the recognition accuracy increases firstly and then keeps stable. Moreover, the recognition accuracy of cLBP is obviously higher than that of cLBP-RGB and cLBP-RSS, the recognition accuracy of cLBP-RSS is higher than that of cLBP-RGB on all of the three datasets. This verifies the following two issues that correspond to our main contributions in this paper: 1) image represented in RSS color space provides more discriminative information for color image recognition than in traditional RGB color space; 2) there has feature redundancy in cLBP, and the proposed color-related patterns learning step can refine the discriminative patterns for representing the image's color information efficiently, and avoid the curse of dimensionality in color image recognition. Moreover, it is worth noting that, in Fig. 6, the recognition accuracy of cLBP increases very slowly or even decreases when the dimension of feature is up to 900.

\begin{figure}[tbp]
\centering
\includegraphics[width=\columnwidth]{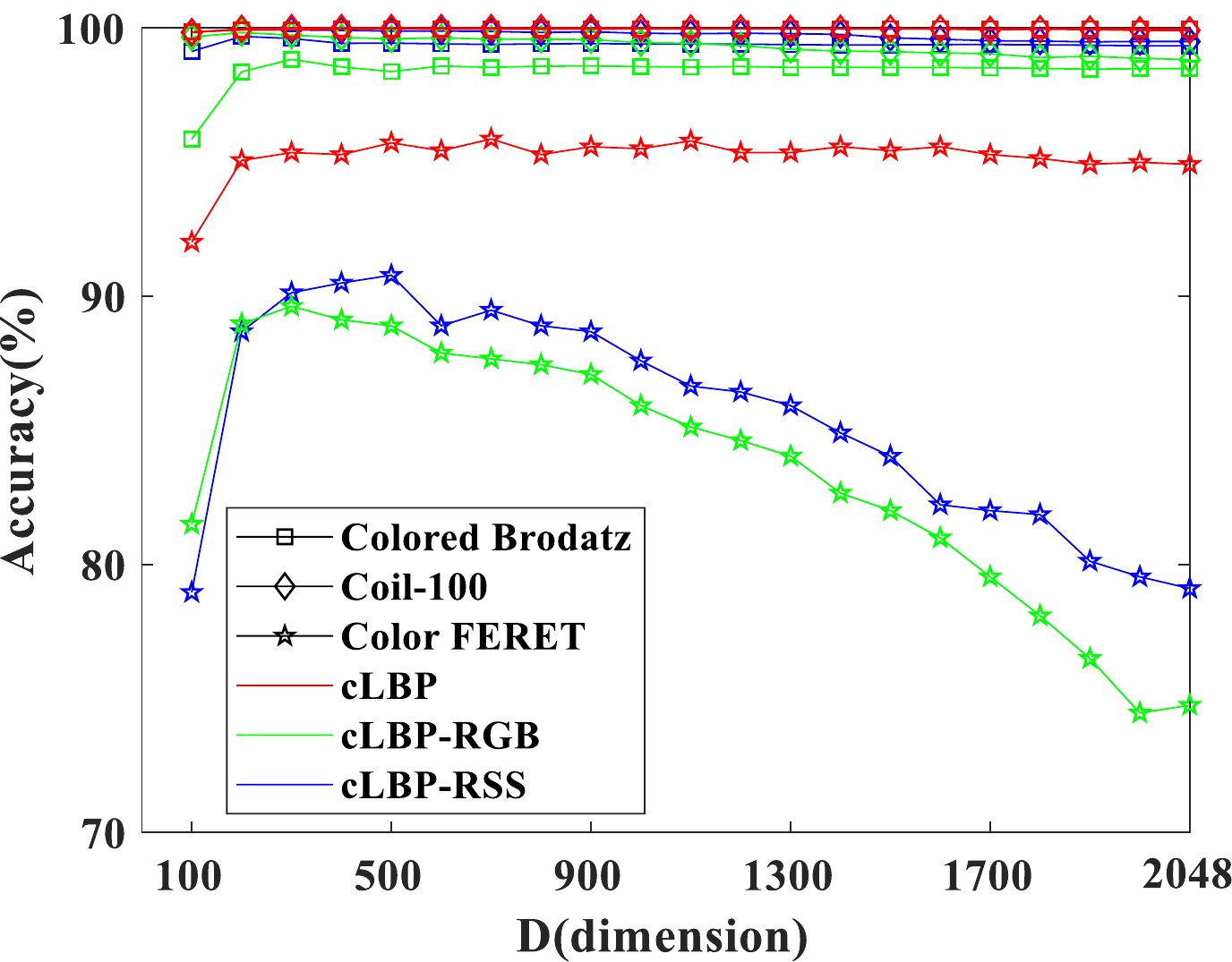}
\caption{The comparison of recognition accuracy by the proposed cLBP-RSS, cLBP-RGB and cLBP with different dimensions of features.}
\label{fig_sim}
\end{figure}
\begin{table}[tbp] \footnotesize
  \centering
  \captionsetup{justification=centering}
  \caption{Recognition accuracy (\%) of color texture images by different methods}
  \begin{spacing}{1.19}
    \setlength{\tabcolsep}{0.1mm}{
    \begin{tabular}{p{6em}<{\centering}p{4em}<{\centering}cccc}
    \toprule
    \multirow{2}[4]{*}{Category} & \multicolumn{4}{c}{Datasets} & \multirow{2}[4]{*}{Average} \\
\cmidrule{2-5}    \multicolumn{1}{c}{} & \multicolumn{1}{p{3.3em}}{No. of features} & \multicolumn{1}{p{4.19em}}{Colored Brodatz} & \multicolumn{1}{p{4.19em}}{STex-512S} & \multicolumn{1}{p{4.19em}}{KTH-TIPS} & \multicolumn{1}{c}{} \\
    \midrule
    \midrule
    QLRBP & 768   & 95.35 & 42.07 & 96.67 & \multicolumn{1}{c}{78.03} \\
    RGB-OC-LBP & 96    & 91.79 & 54.06 & 95.4  & \multicolumn{1}{c}{80.42} \\
    LBP   & 768   & 91.86 & 66.33 & 97.78 & \multicolumn{1}{c}{85.32} \\
    LCVBP & 237   & 91.88 & 66.18 & 97.89 & \multicolumn{1}{c}{85.32} \\
    LBPC  & 542   & 99.67 & 70.72 & 98.64 & \multicolumn{1}{c}{89.68} \\
    MDLBP & 2048  & 98.49 & 69.79 & 97.53 & \multicolumn{1}{c}{88.6} \\
    \midrule
    cLBP($D$=100) & 100   & 99.86 & 69.83 & 99.62 & \multicolumn{1}{c}{89.77} \\
    cLBP($D$=400) & 400   & 99.96 & 86.26 & 99.88 & \multicolumn{1}{c}{95.37} \\
    cLBP($D$=900) & 900   & \textbf{99.97} & \textbf{89.82} & \textbf{100} & \textbf{96.6} \\
    \bottomrule
    \end{tabular}}%
  \end{spacing}
  \label{tab:addlabel}%
\end{table}%

\begin{table}[tbp] \footnotesize
  \centering
  \captionsetup{justification=centering}
  \caption{Recognition accuracy (\%) of color object images by different methods}
    \begin{spacing}{1.19}
    \setlength{\tabcolsep}{0.1mm}{
    \begin{tabular}{p{6em}<{\centering}p{4em}<{\centering}ccccc}
    \toprule
    \multirow{2}[3]{*}{Category} & \multicolumn{5}{c}{Datasets} & \multicolumn{1}{c}{\multirow{2}[3]{*}{Average}} \\
\cmidrule{2-6} \multicolumn{1}{c}{} & \multicolumn{1}{p{3.3em}}{No. of features}  & \multicolumn{1}{p{5em}}{Wang or SIMPLIcity} & \multicolumn{1}{p{4.19em}}{Corel-10k} & \multicolumn{1}{p{4.19em}}{Coil-100} & \multicolumn{1}{p{4.19em}}{FTVL} &  \\
    \midrule
    \midrule
    QLRBP & 768 & 89    & 55.61 & 98.21 & 96.69 & 84.88 \\
    RGB-OC-LBP & 96 & 83.3  & 54.88 & 99.64 & 96.13 & 83.48 \\
    LBP   & 768 & 87.1  & 58.33 & 98.61 & 94.83 & 84.72 \\
    LCVBP & 237 & 85.9  & 59.33 & 98.96 & 97.57 & 85.44 \\
    LBPC  & 542 & 89.9  & 63.71 & 99.82 & 97.72 & 87.79 \\
    MDLBP & 2048 & 90.4  & 62.51 & 99.5  & 98.86 & 87.82 \\
    \midrule
    cLBP($D$=100) & 100 & 89.5  & 64.25 & 99.83 & 98.71 & 88.07 \\
    cLBP($D$=400) & 400 & 90.9  & 72.64 & 100   & 99.62 & 90.79 \\
    cLBP($D$=900) & 900 & \textbf{92.2} & \textbf{74.66} & \textbf{100} & \textbf{99.81} & \textbf{91.67} \\
    \bottomrule
    \end{tabular}}%
    \end{spacing}
  \label{tab:addlabel}%
\end{table}%

\begin{table}[tbp] \footnotesize
  \centering
  \captionsetup{justification=centering}
  \caption{Recognition accuracy (\%) of color face images by different methods}
    \begin{spacing}{1.19}
    \setlength{\tabcolsep}{0.1mm}{
    \begin{tabular}{p{8em}<{\centering}p{4em}<{\centering}ccc}
    \toprule
    \multirow{2}[4]{*}{Category} & \multicolumn{3}{c}{Datasets} & \multicolumn{1}{c}{\multirow{2}[4]{*}{Average}} \\
\cmidrule{2-4}    \multicolumn{1}{c}{} & \multicolumn{1}{p{3.3em}}{No. of features} & \multicolumn{1}{p{4.19em}}{Color FERET} & \multicolumn{1}{p{4.19em}}{AR face} &  \\
    \midrule
    \midrule
    QLRBP & 768$\times$4  & 84.54 & 63.86 & 74.2 \\
    RGB-OC-LBP & 96$\times$4  & 89.33 & 92.00    & 90.67 \\
    LBP   & 768 $\times$4   & 75.83 & 88.29 & 82.06 \\
    LCVBP & 237 $\times$4  & 87.15 & 91.36 & 89.26 \\
    LBPC  & 542 $\times$4   & 89.99 & 87.71 & 88.85 \\
    MDLBP & 2048$\times$4  & 79.17 & 81.36 & 80.27 \\
    \midrule
    cLBP($D$=100) & 100$\times$4   & 92.02 & 77.5  & 84.76 \\
    cLBP($D$=400) & 400$\times$4  & 95.28 & 92.14 & 93.71 \\
    cLBP($D$=900) & 900$\times$4  & \textbf{95.94} & \textbf{97.21} & \textbf{96.58} \\
    \bottomrule
    \end{tabular}}%
    \end{spacing}
  \label{tab:addlabel}%
\end{table}%
\subsection{Comparison With Existing Methods}
In this subsection, the proposed cLBP is compared with six LBP variants mentioned in section \uppercase\expandafter{\romannumeral3}.A. For fairly comparison, the dimension of learned features in cLBP is selected as $D$=100, 400 and 900 respectively. The experiments on color texture, object and face image recognition are designed to evaluate the performance of the propose cLBP comprehensively.

Firstly, we conduct the color texture image recognition on ``STex-512S", ``Colored Brodatz" and ``KTH-TIPS" image datasets by the proposed cLBP and other LBP variants. Table \uppercase\expandafter{\romannumeral2} shows the recognition accuracy on those three color texture image datasets by all the comparison methods. From this table, we can find that the cLBP with $D$=100 achieves 9.35\% higher in average recognition accuracy than RGB-OC-LBP which has almost the same dimension of feature, and 11.74\% higher in average recognition accuracy than QLRBP which has more than seven times number of features. When $D$ is up to 900, the recognition accuracy of cLBP is obviously higher than all other methods. It achieves almost 20\% higher in recognition accuracy on ``STex-512S" than LBPC which has achieved the highest recognition accuracy among the state of the art methods.

Secondly, we test the color object recognition ability of the proposed cLBP and comparison methods on ``Wang or SIMPLIcity", ``Corel-10k", ``FTVL" and ``Coil-100" datasets. The recognition accuracy is shown in Table \uppercase\expandafter{\romannumeral3}. It can be seen from this table that the average recognition accuracy of cLBP with $D$=100 is higher than all other methods. When $D$ is up to 900, the recognition of our proposed method on ``Corel-10k" dataset is up to 74.66\%, while the highest recognition accuracy of the state of the art methods on this dataset is only 63.71\%.

Thirdly, the color face recognition ability of all comparison methods are evaluated on the ``Color FERET" and ``AR face" datasets. In this experiment, each face image in the datasets is divided into $2\times2$ sub-regions, and the feature vectors of all these four sub-regions are concatenated as the final feature vector for classification. Table \uppercase\expandafter{\romannumeral4} shows the recognition accuracy achieved by the proposed cLBP and all comparison methods. From the table, it can be found that the proposed cLBP also achieve higher recognition accuracy than other methods especially when the feature dimension of cLBP is $D$=900.

From the above three experiments, it can be concluded that the proposed cLBP has an excellent performance in color images recognition with low dimension of features. The combination of RSS and RGB spaces in cLBP provides more discriminative patterns than other methods in color image recognition. Moreover, the learning strategy in the proposed method is effective in selecting the dominant color-related patterns in color image recognition.
\begin{figure*}[!ht]
\centering
\includegraphics[width=\textwidth]{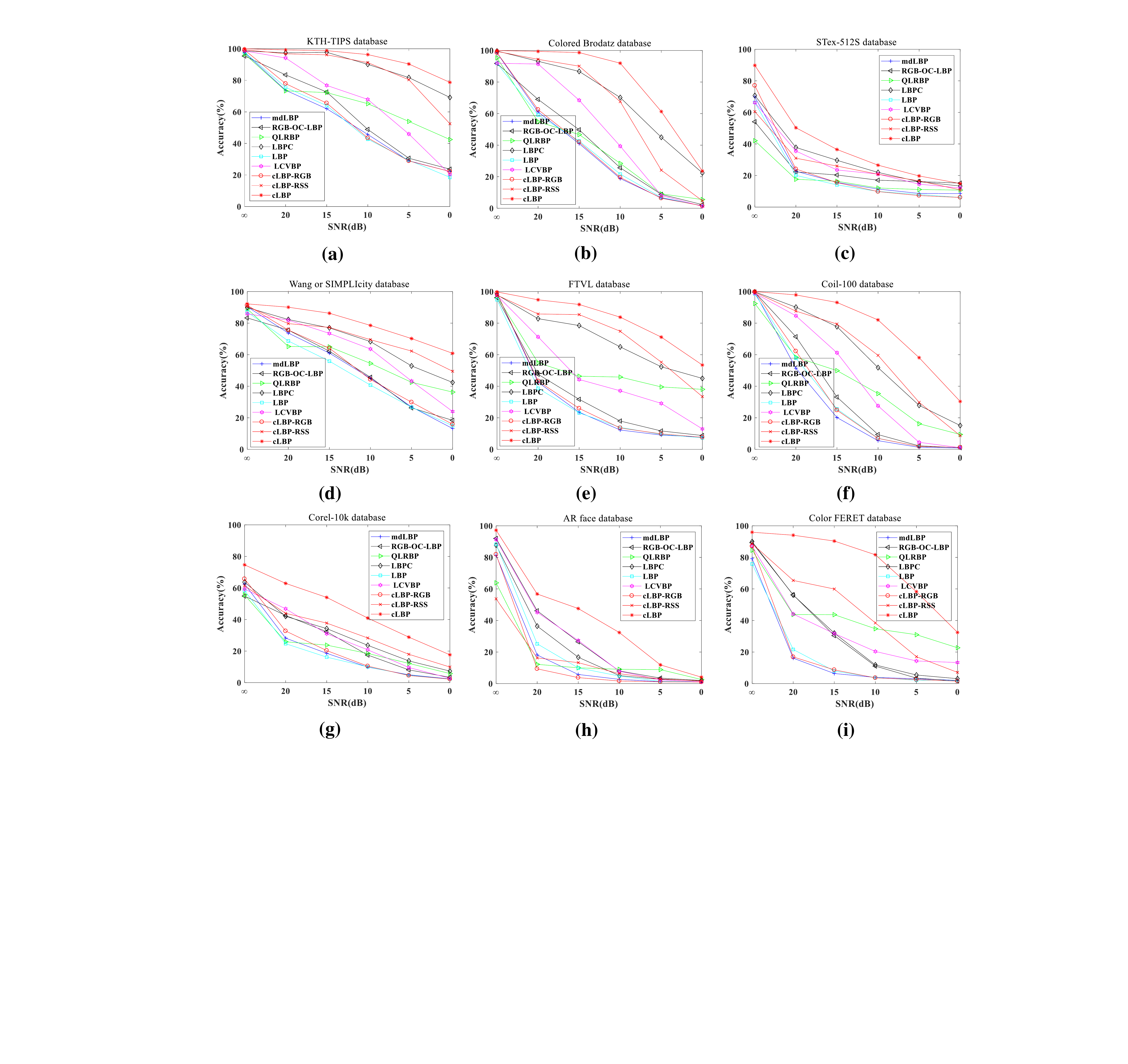}
\caption{Comparison of image recognition accuracy by the proposed cLBP and other methods under different noisy conditions.}
\label{fig_sim}
\end{figure*}
\subsection{Noise Robustness of the cLBP}

As mentioned above, the calculation process of RSS space can offset the interference of noise. This experiment is designed to evaluate the color image recognition ability by the proposed cLBP and other comparison methods under noisy condition. In the testing procedure of all methods, the testing images are corrupted by Gaussian noise with signal-to-noise ratios (SNR) varying from 20 dB to 0 dB with 5 decrements, and the dimension of feature in cLBP is set as $D$=900. Fig. 7 shows the recognition accuracy achieved by different methods on the first three groups (nine) image datasets. It can be found that the proposed cLBP has absolutely higher recognition accuracy than other LBP variants under different noisy condition. When the SNR is decreased to 0 dB (i.e., half noise and half signal), the recognition accuracy of the proposed cLBP on ``KTH-TIPS" image dataset is still up to 80\%. In order to further verify the noise robustness of the proposed cLBP mainly contributed from whether the RSS space or RGB space, the cLBP-RGB ($D$=900) and cLBP-RSS ($D$=900) used in Section \uppercase\expandafter{\romannumeral3}. C are also compared in this experiment, and the recognition accuracy of these two methods on all image datasets is also provided in Fig. 7. It can be found that the cLBP-RSS has higher recognition accuracy than cLBP-RGB and lower than cLBP on all of the nine image datasets. This verifies the contribution of our proposed method that the RSS space provides high noise robustness in cLBP for image recognition.

\subsection{Illumination invariance of cLBP}
In the experiments designed in Section \uppercase\expandafter{\romannumeral3}. D, some images in the ``KTH-TIPS", ``color FERET" and ``AR face" image datasets are obtained under illumination variation. Therefore, the experimental results shown in tables \uppercase\expandafter{\romannumeral2}  and \uppercase\expandafter{\romannumeral4} have preliminary verified the effectiveness of the proposed cLBP in color image recognition under illumination variation condition. In this experiment, three color image datasets, i.e., ``Outex-14" \cite{ojala2002outex}, ``ALOI" \cite{geusebroek2005amsterdam} and ``CUReT" \cite{dana1999reflectance} which are specifically constructed for illumination invariant image recognition as well as have been widely used in existing methods, are utilized for further validating the recognition ability of the proposed cLBP under illumination variation. The dimension of feature in cLBP is fixed to $D$=900. Table \uppercase\expandafter{\romannumeral5} shows the recognition accuracy achieved on ``Outex-14", ``ALOI" and ``CUReT" image datasets by different methods. It can be found that, the recognition accuracy on all of three image datasets by cLBP is significantly higher than other methods. This experimental results is consistent with the property we analyzed in Section \uppercase\expandafter{\romannumeral2}. A that the color images represented in RSS space are difficult to be affected by illumination variation.
\begin{table}[tbp] \footnotesize
  \centering
  \captionsetup{justification=centering}
  \caption{Recognition accuracy (\%) on the ``Outex-14", ``ALOI" and ``CUReT" datasets by different methods}
  \begin{spacing}{1.19}
    \setlength{\tabcolsep}{0.1mm}{
    \begin{tabular}{p{6em}<{\centering}p{4em}<{\centering}cccc}
    \toprule
    \multirow{2}[4]{*}{Category} & \multicolumn{4}{c}{Datasets} & \multirow{2}[4]{*}{Average} \\
\cmidrule{2-5}    \multicolumn{1}{c}{} & \multicolumn{1}{p{3.3em}}{No. of features} & \multicolumn{1}{p{4.19em}}{Outex-14} & \multicolumn{1}{p{4.19em}}{ALOI} & \multicolumn{1}{p{4.19em}}{CUReT} & \multicolumn{1}{c}{} \\
    \midrule
    \midrule
    QLRBP & 768   & 66.49 & 75.68 & 93.67 & \multicolumn{1}{c}{78.61} \\
    RGB-OC-LBP & 96    & 90.08 & 94.33 & 95.81  & \multicolumn{1}{c}{93.41} \\
    LBP   & 768   & 85.83 & 91.7 & 97.72 & \multicolumn{1}{c}{91.75} \\
    LCVBP & 237   & 85.42 & 86.88 & 98.01 & \multicolumn{1}{c}{90.1} \\
    LBPC  & 542   & 85.15 & 78.78 & 98.52 & \multicolumn{1}{c}{87.48} \\
    MDLBP & 2048  & 83.36 & 87.13 & 94.53 & \multicolumn{1}{c}{88.34} \\
    \midrule
    cLBP & 900   & \textbf{93.41} & \textbf{96.06} & \textbf{99.59} & \textbf{96.35} \\
    \bottomrule
    \end{tabular}}%
  \end{spacing}
  \label{tab:addlabel}%
\end{table}%

\section{Conclusion}

In this paper, a color-related local binary pattern (cLBP) which learns the dominant patterns from the decoded LBP was proposed for color images recognition. In the proposed method, the relative similarity space (RSS) is firstly introduced to obtain the color similarity between the three channels of color images. Theoretical analysis show that the RSS space provides more discriminative information, has higher noise robustness and has the property of illumination invariance compared with the traditional RGB space. Secondly, the decoded LBP is employed to describe the color image on the combination of the RSS and RGB color space. The decoded LBP provides an excellent way to mine the correlation information between the LBP feature maps corresponds to each color channel. Since the dimension of decoded LBP is too high to easily cause the curse of dimensionality in classification, thirdly, a feature learning strategy is used to learn the dominant color-related patterns to reduce the dimension of feature vector and further improve the recognition accuracy. Finally, the proposed cLBP is compared with six state of the art LBP variants which are specifically designed for color image recognition. The experimental results conducted on the color texture, object and face image recognition show that, the proposed cLBP achieved obviously higher recognition accuracy than other methods and has low dimension of features under noise-free, noisy and illumination variation conditions.

Although the learning strategy employed in the proposed method can reduce the dimension of features and learn the dominant patterns to improve the recognition accuracy, the dimension of features are manually fixed by experimental results, how to automatically estimate the dimension of features in the proposed method is the future work.

\ifCLASSOPTIONcaptionsoff
  \newpage
\fi



\bibliographystyle{IEEEtran}
\bibliography{ref}
%
%
%

%

\begin{IEEEbiography}{Bin Xiao}
received his B.S. and M.S. degrees in Electrical Engineering from Shanxi Normal University, Xian, China in 2004 and 2007, received his Ph. D. degree in computer science from Xidian University, XiAn, China in 2012. He is now a professor at Chongqing University of Posts and Telecommunications, Chongqing, China. His research interests include image processing and pattern recognition.
\end{IEEEbiography}

\vspace{-20 mm}
\begin{IEEEbiography}{Tao Geng}
was born in 1998. He is a postgraduate student with the School of Computer Science and Technology, Chongqing University of Posts and Telecommunications, Chongqing, China. His research interest is image recognition.
\end{IEEEbiography}

\vspace{-20 mm}
\begin{IEEEbiography}{Xiuli Bi}
received her B.Sc. and M.Sc. degrees from Shanxi Normal University, China, in 2004 and 2007 respectively, and Ph.D. degree in Computer Science from the University of Macau in 2017. She is currently an associate professor at the College of Computer Science and Technology, Chongqing University of Posts and Telecommunications, China. Her research interests include medical image processing; multimedia security and image forensics.
\end{IEEEbiography}

\vspace{-20 mm}
\begin{IEEEbiography}{Weisheng Li}
received his B.S. degree from School of Electronics and Mechanical Engineering at Xidian University, Xian, China in July 1997. He received his M.S. degree and Ph.D. degree from School of Electronics and Mechanical Engineering and School of Computer Science and Technology at Xidian University in July 2000 and July 2004, respectively. Currently he is a professor of Chongqing University of Posts and Telecommunications. His research focuses on intelligent information processing and pattern recognition.
\end{IEEEbiography}




\end{document}